# Inference with Separately Specified Sets of Probabilities in Credal Networks

**José Carlos Ferreira da Rocha**
Escola Politcnica, Univ. de São Paulo
and Univ. Estadual de Ponta Grossa
Ponta Grossa, PR Brazil
jrocha@uepg.br

**Fabio Gagliardi Cozman**
Escola Politécnica, Univ. de São Paulo
São Paulo, SP Brazil
fgcozman@usp.br

## Abstract

We present new algorithms for inference in *credal networks* — directed acyclic graphs associated with sets of probabilities. Credal networks are here interpreted as encoding *strong* independence relations among variables. We first present a theory of credal networks based on separately specified sets of probabilities. We also show that inference with polytrees is NP-hard in this setting. We then introduce new techniques that reduce the computational effort demanded by inference, particularly in polytrees, by exploring separability of credal sets.

## 1 INTRODUCTION

In this paper we deal with models that represent uncertainty through directed acyclic graphs and sets of probability measures. We refer to such sets as *credal sets* [18], and to such graphical structures as *credal networks* [7, 13]. Credal sets have been used to represent imprecision in probability statements, to summarize opinions of groups of experts, and to conduct robustness analysis of probabilistic models [11, 12]. Overall, credal sets offer a very general and flexible tool for uncertainty representation.

Our goal is to produce *inferences*, here understood as tight lower/upper bounds on posterior probabilities given evidence. We adopt the concept of *strong independence* in our credal networks [6, 8]. Inference with strong independence relations is quite demanding computationally because it involves non-linear optimization problems.

In Section 2 we present the theory of credal sets and introduce new concepts that are necessary to properly handle *separately specified* credal sets. We describe a new theory of credal networks with separately specified credal sets and review existing inference methods in Sections 3 and 4 respectively. We prove a result of independent interest: in polytrees, exact inference is NP-hard in the presence of credal sets. We then investigate the use of redundancy elimination algorithms in credal networks, an idea that has been suggested several times but that has never been empirically tested (Section 5).

In Section 6 we present a new approach to separately specified credal sets and discuss techniques that greatly reduce the computational effort demanded by inference. Our techniques explore separability of credal sets in a general and profitable manner. In particular, we show that inference with polytrees is viable even in situations previously thought to be unmanageably complex. We also compare our algorithm to the message-based 2U algorithm, and discuss the generality of our approach.

The results presented suggest that exact inferences can be produced for medium-sized credal networks, particularly polytrees. This assertion is significantly more optimistic than existing results in the literature, which seem to suggest that credal networks can only be handled in practice by approximate methods.

## 2 CREDAL SETS AND CONDITIONAL INDEPENDENCE

A *credal set* $K$ is a set of probability measures.[1] Credal sets generalize probability intervals and belief functions, offering a general framework for the representation of imprecision and indeterminacy about probability [22].

A credal set defined by a collection of probability densities $p(X)$ is denoted by $K(X)$. Here we focus only on *closed convex* sets of probability densities. Given a credal set $K(X)$ and a function $f(X)$, the *lower* and *upper* expectations of $f(X)$ are defined respectively as $\underline{E}[f(X)] = \min_{p(X) \in K(X)} E_p[f(X)]$ and $\overline{E}[f(X)] =$ $\max_{p(X) \in K(X)} E_p[f(X)]$ (here $E_p[f(X)]$ indicates standard expectation). Similarly, the *lower probability* and the *upper probability* of event $A$ are defined respectively as $\underline{P}(A) = \min_{p(X) \in K(X)} P(A)$ and $\overline{P}(A) = \max_{p(X) \in K(X)} P(A)$. A set of probability measures, its convex hull, and its vertices produce the *same* lower/upper expectations and lower/upper probabilities. Note that a "vertex" here means a probability distribution.

---

[1]The first schemes proposed in the literature to refer to general credal sets as objects of intrinsic interest seem to be from Levi, who used $B$ [18], and from Giron and Rios, who used $K$ [14] — we opted for the latter merely because its sound suggests the word "credal." Other notations employed for credal sets are $\mathcal{M}$, $\mathcal{P}$, $\mathcal{C}$, $\Pi$, $\Gamma$ (the last one is often used in robust statistics).

Conditioning is equated to elementwise application of Bayes rule in a credal set; the posterior credal set is the union of all posterior probability measures [22]. A *conditional* credal set $K(X|Y = y)$ contains conditional probability densities $p(X|Y = y)$ for random variables $X$ and $Y$.

There are two different ways to represent credal sets conditioned on random variables. First, consider the collection of *separately specified* conditional credal sets $K(X|Y = y)$, one credal set for each value of $Y$, which we denote by $K(X|Y)$. Second, consider the direct specification of a set of functions $p(X|Y)$ — we call such a set an *extensive* conditional credal set, and denote it by $L(X|Y)$.[2] It turns out that the differences between $K(X|Y)$ and $L(X|Y)$ are critical to the results in this paper, but these differences have not been emphasized previously in the literature.

The sets $L(X|Y)$ might seem to carry all conditional information for $X$ conditional on $Y$, but this is not correct. Consider the next example, where a set $L^*(X, Y|Z)$ induces a joint credal set $K^*(X, Y, Z)$ but does not capture all the conditional distributions $p(X, Y|Z)$ generated from $K^*(X, Y, Z)$.

**Example 1** Take four functions over binary variables $X$ and $Y$; these functions are defined by vectors of the form $[f(x_0, y_0), f(x_0, y_1), f(x_1, y_0), f(x_1, y_1)]$. The functions are: $f_1(X, Y) = [0.14, 0.06, 0.56, 0.24]$, $f_2(X, Y) = [0.48, 0.32, 0.12, 0.08]$, $f_3(X, Y) = [0.32, 0.08, 0.48, 0.12]$, and $f_4(X, Y) = [0.63, 0.27, 0.07, 0.03]$. Consider a binary variable $Z$ and construct two functions over $(X, Y, Z)$: $l_1(X, Y|z_0) = f_1(X, Y)$; $l_1(X, Y|z_1) = f_2(X, Y)$; $l_2(X, Y|z_0) = f_3(X, Y)$; $l_2(X, Y|z_1) = f_4(X, Y)$. Consider now a set $L^*(X, Y|Z)$ with extreme points $l_1(X, Y|Z)$ and $l_2(X, Y|Z)$. Define distributions $p_1(Z)$ and $p_2(Z)$ by $p_1(Z = z_0) = 0.4$ and $p_2(Z = z_0) = 0.6$. Finally, suppose the joint credal set $K^*(X, Y, Z)$ contains four extreme distributions:

[2]Moral and Cano use the term *conditioned to the elements* to indicate separately specified credal sets (the latter term is used by Walley [22] and Cozman [7]). Moral and Cano use the term *global information* to indicate extensive credal sets [19]. We prefer to reserve the word "information" to its technical meaning in information theory, and while the term "global conditional credal set" could be used for $L(X|Y)$, the word "global" might be misunderstood as to mean a set with joint distributions. We have thus decided to adopt the term "extensive" here.

$l_1(X, Y|Z)p_1(Z)$, $l_1(X, Y|Z)p_2(Z)$, $l_2(X, Y|Z)p_1(Z)$, and $l_2(X, Y|Z)p_2(Z)$. Even though $K^*(X, Y, Z)$ was directly built from $L^*(X, Y|Z)$, the set of all conditional distributions computed from $K^*(X, Y, Z)$ is *larger* than $L^*(X, Y|Z)$! For example, if we take the convex combination $0.4 l_1(X, Y|Z)p_1(Z) + 0.6 l_2(X, Y|Z)p_2(Z)$, we obtain $p(X, Y|Z) = l_3(X, Y|Z)$ where $l_3(X, Y|z_0) = (4f_1(X, Y) + 9f_3(X, Y))/13$; $l_3(X, Y|z_1) = (f_2(X, Y) + f_4(X, Y))/2$. Note that $l_3(X, Y|Z)$ is not a convex combination of $l_1(X, Y|Z)$ and $l_2(X, Y|Z)$.

Consider now the concepts of independence and conditional independence. In this paper we focus on the concept of independence that is most often adopted for credal sets: Variables $X$ and $Y$ are *strongly independent* when every vertex of $K(X, Y)$ satisfies stochastic independence of $X$ and $Y$ — every vertex satisfies $p(X|Y) = p(X)$ and $p(Y|X) = p(Y)$ for possible values of $(X, Y)$ [6, 8, 10].

We adopt the following definition for *conditional* strong independence:

**Definition 1** *Variables $X$ and $Y$ are strongly independent conditional on variable $Z$ when the sets $K(X, Y|Z = z)$ have vertices satisfying stochastic independence of $X$ and $Y$ for every value of $Z$.*

For conditions that imply strong independence, and that can be adapted to obtain Definition 1, the reader is referred to [8]. Strong independence is quite reasonable in a "sensitivity analysis" interpretation of credal sets, where credal sets are viewed as containing a "correct" model. Several other concepts of conditional independence are discussed by Moral and Cano [19].

To dramatically illustrate the difference between the sets $K(X|Y)$ and $L(X|Y)$, consider an alternative definition for conditional independence:

**Definition 2** *Variables $X$ and $Y$ are strongly independent conditional on variable $Z$ when every vertex of $L(X, Y|Z)$ satisfies stochastic independence of $X$ and $Y$ conditional on $Z$ (for every value of $Z$).*

If $X$, $Y$ and $Z$ satisfy Definition 2, they immediately satisfy Definition 1; the reverse is not always true:

**Example 2** Consider three binary variables $X$, $Y$ and $Z$ and construct a set $L(X, Y|Z)$ with three vertices $l_1(X, Y|Z)$, $l_2(X, Y|Z)$ and $l_3(X, Y|Z)$ as defined in Example 1. The set $L(X, Y|Z)$ induces a collection of sets $K(X, Y|Z)$. Note that $X$ and $Y$ are not independent conditional on $Z$ according to Definition 2 (because $l_3(X, Y|Z)$ is a vertex of $L(X, Y|Z)$ and does not factorize

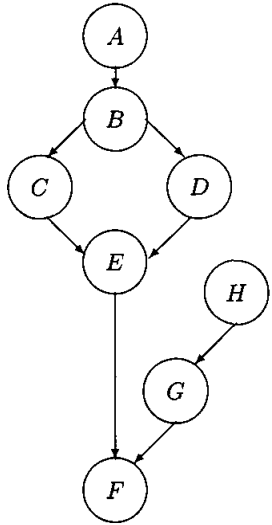

$P(a_0) \in [1/2, 3/5]$
$P(b_0|a_0) \in [1/2, 3/5]$
$P(b_0|a_1) \in [2/5, 1/2]$
$P(d_0|b_0) \in [1/5, 4/5]$
$P(d_0|b_1) \in [1/10, 1/2]$
$P(e_0|c_0, d_0) \in [1/10, 9/10]$
$P(e_0|c_0, d_1) = 1/2$
$P(e_0|c_1, d_0) \in [3/10, 1/2]$
$P(e_0|c_1, d_1) \in [1/10, 1/2]$
$P(e_0|c_2, d_0) = 1/2$
$P(e_0|c_2, d_1) \in [1/2, 3/5]$
$P(f_0|e_0, g_0) \in [1/5, 51/100]$
$P(f_0|e_0, g_1) \in [1/4, 1/2]$
$P(f_0|e_1, g_0) \in [9/20, 11/20]$
$P(f_0|e_1, g_1) \in [7/20, 11/20]$
$P(h_0) \in [1/5, 11/20]$
$P(g_0|h_0) \in [1/5, 1/2]$
$P(g_0|h_1) \in [3/10, 1/2]$
$K(C|b_0) = \{[1/3, 1/3, 1/3], [7/10, 1/10, 1/5]\}$
$K(C|b_1) = \{[1/3, 1/3, 1/3], [1/2, 1/2, 3/10]\}$

Figure 1: Example credal network: all variables are binary, except $C$, which is ternary.

when restricted to $\{Z = z_0\}$), but $X$ and $Y$ are independent conditional on $Z$ according to Definition 1 (because $l_3(X, Y|Z)$ does not produce vertices of $K(X, Y|Z)$).

Examples 1 and 2 show that Definition 2 can lead to somewhat paradoxical situations. For example, we might take variables $X$ and $Y$ to be independent conditional on $Z$ and then, because of "unforeseen" vertices of $L(X, Y|Z)$, $X$ and $Y$ are actually *dependent* conditional on a fixed value of $Z$ (according to Definition 1).

## 3 CREDAL NETWORKS WITH SEPARATELY SPECIFIED CREDAL SETS

To represent a set of joint probability distributions over a large number of variables, we can resort to credal networks — an obvious generalization of the successful *Bayesian network* model [20]. To review, a Bayesian network is based on a directed acyclic graph and a collection of variables **X**. Each node in the graph is associated with a variable $X_i$ and with a conditional density $p(X_i|\Pi_i)$, where $\Pi_i$ indicates the parents of $X_i$ in the graph. Every Bayesian network satisfies the *Markov condition*: every variable is independent of its nondescendants nonparents conditional on its parents.

We define a credal network over variables **X** using a directed acyclic graph where every node is associated with a variable $X_i$, and each variable is associated with credal sets that contain the probability measures for $X_i$ conditional on $\Pi_i$. We assume in this paper that every variable is *strongly* *independent* of its nondescendants nonparents conditional on its parents.

Figure 1 shows a credal network. In this example, note that the credal sets for a variable $X_i$ are specified separately for each value of $\Pi_i$. This is a common situation: no credal network in the literature is specified directly through sets $L(X_i|\Pi_i)$. There are in fact good reasons to avoid direct specification of sets $L(X_i|\Pi_i)$ when building a credal network. First, functions is these sets are not normalized, and it is difficult to express beliefs by manipulating these functions directly. Second, the use of $L(X_i|\Pi_i)$ may induce unforeseen vertices into the joint credal set $K(\mathbf{X})$ when the network is taken in its entirety (as in Example 1).

Even though credal networks always use separately specified credal sets, algorithms in the literature do not use this fact explicitly to simplify computations. To our knowledge, the present paper contains the first general exact algorithms that explicitly deal with separability of credal sets for inference (Section 6).

Once the sets $K(X_i|\Pi_i)$ for all variables in a credal network are specified, we must indicate how the various sets are to be combined as we vary the values of $\Pi_i$.

**Example 3** Consider Figure 1. If we focus on variable $B$, we could construct at least two sets $K(A, B)$. The first set contains eight vertices, and we obtain these eight vertices by considering every possible extreme value of $P(a_0)$, $P(b_0|a_0)$ and $P(b_0|a_1)$. The second set contains two vertices: one vertex is obtained by choosing $P(a_0) = 1/2$, $P(b_0|a_0) = 1/2$ and $P(b_0|a_1) = 2/5$; the other vertex is obtained by choosing $P(a_0) = 3/5$, $P(b_0|a_0) = 3/5$ and $P(b_0|a_1) = 1/2$.

To be able to describe the construction of joint distributions from sets $K(X|Y)$, it is conveniente to define *concatenation* operations for distributions and credal sets.

**Definition 3** *The concatenation of functions* $p(X|Y = y)$, *denoted by* $\mathsf{CONCAT}_Y(p(X|Y))$, *is the function* $g(X, Y)$ *defined by* $g(X, y) = p(X|Y = y)$; *that is*, $g(X, Y)$ *is formed by "concatenating" the functions* $p(X|Y = y)$.

**Definition 4** *The concatenation of sets* $K(X|Y)$, *denoted by* $\mathsf{CONCAT}_Y(K(X|Y))$, *is the set*

$$\{\mathsf{CONCAT}_Y(p(X|Y = y)) : p(X|Y = y) \in K(X|Y = y)\}.$$

Essentially, when we have a set $K(X|Y)$, we assume that every concatenation of distributions from $K(X|Y = y)$ is valid. This strategy corresponds to choosing eight vertices for $K(A, B)$ in Example 3. Note that this strategy constructs the largest possible sets $L(X|Y)$; by taking the largest possible credal set, we do not introduce any additional constraint into our model.

**Example 4** Consider variable $F$ in Figure 1. There are four possible values for $(E, G)$. The credal set $\text{CONCAT}_{E,G}(K(F|E,G))$ contains $2^4$ vertices.

We are finally prepared to give the central definition regarding the use of credal networks with strong conditional independence:

**Definition 5** *The* strong extension *of a credal network is the convex hull of probability densities that satisfy the Markov condition on the network:*

$$K(\mathbf{X}) = \text{CONVEXHULL}\left(\begin{array}{c}\prod_i p(X_i|\Pi_i) : \\ p(X_i|\Pi_i) \in L(X_i|\Pi_i)\end{array}\right). \tag{1}$$

*If $L(X_i|\Pi_i)$ is equal to $\text{CONCAT}_{\Pi_i}(K(X_i|\Pi_i))$, we say the strong extension is* separately specified.

The concept of strong extension is not new [5, 8, 21], but we are careful in Definition 5 to emphasize the possible ways to specify $L(X_i|\Pi_i)$, a point that is algorithmically important and has been neglected previously.

In this paper we assume that variables are discrete with finitely many values, and that local sets $K(X_i|\Pi_i)$ are specified by a finite number of vertices. Even with these simplifying assumptions, it should be clear that separately specified strong extension are extremely complex objects. A very simple credal network can produce a strong extension with a huge number of vertices:

**Example 5** Consider a network with four variables $X$, $Y$, $Z$ and $W$; $W$ is the sole child of $X$, $Y$ and $Z$. Suppose that all variables have three values and that every local credal set has only three vertices. The number of possible combinations of vertices of local credal sets is $3^{30}$.

Our purpose is to compute the lower probability $\underline{P}(X_q|\mathbf{X}_E)$ for values of the *query* variable $X_q$ conditional on evidence $\mathbf{X}_E$. We assume that $X_q$ is not in $\mathbf{X}_E$ and that $\underline{P}(\mathbf{X}_E) > 0$. We deal with a non-linear optimization problem, because the objective function is

$$\sum_{\mathbf{X}\setminus\{X_q,\mathbf{X}_E\}} p(\mathbf{X}) / \sum_{\mathbf{X}\setminus\{\mathbf{X}_E\}} p(\mathbf{X}), \tag{2}$$

which, given independence relations, is equal to

$$\frac{\sum_{\mathbf{X}\setminus\{X_q,\mathbf{X}_E\}} \prod_i p(X_i|\Pi_i)}{\sum_{\mathbf{X}\setminus\{\mathbf{X}_E\}} \prod_i p(X_i|\Pi_i)}, \tag{3}$$

subject to the constraint that $p(X_i|\Pi_i) \in L(X_i|\Pi_i)$ for all $X_i$. It is known that the maximum of (3) is attained at a vertex of the strong extension (1) [7, 13].

Strong extensions generalize Bayesian networks; consequently, we cannot expect to find polynomial algorithms for inferences with strong extensions. In Bayesian networks we can find polynomial algorithms for some special structures, for example, polytrees. Unfortunately, strong extensions defy polynomial inference even for polytrees:

**Theorem 1** *Calculation of lower/upper probabilities in strong extensions defined using polytrees is a NP-complete problem.*

The proof is in Appendix A.

## 4 EXISTING INFERENCE ALGORITHMS

Inferences can be generated for strong extensions by enumerating all vertices indicated by Expression (1). Note that several credal sets in a network may be irrelevant for a particular inference, due to d-separation properties [7]; we assume that such irrelevant credal sets are *always* discarded before inference starts.

To organize the enumeration of vertices in a credal network, Cano, Cano and Moral suggest a transformation that takes the original credal network and transforms the network into a standard Bayesian network [2]. The CCM transform adds a *transparent* variable for each local credal set in the network: If $X_i$ is associated with $L(X_i|\Pi_i)$, then a transparent variable $X_i'$ is added as a parent of $X_i$ and a single distribution $p(X_i|X_i',\Pi_i)$ replaces $L(X_i|\Pi_i)$. The number of values for $X_i'$ is equal to the number of vertices of $L(X_i|\Pi_i)$, so that each instantiation of $X_i'$ is equivalent to choosing one of these vertices. After the CCM transform is applied, inference can be performed using any algorithm for Bayesian networks. This "extensive" CCM transform has been implemented in the JavaBayes system (a freely available inference engine distributed at http://www.cs.cmu.edu/javabayes).

To process separately specified credal sets, Moral and Cano suggest that the CCM transform should be modified so as to add a transparent variable *for each* value of $\Pi_i$ [4]. This seems to be the only work, previous to the present one, that has paid attention to the difference between separately specified and extensive credal sets. Moral and Cano derive a number of approximate algorithms based on probability trees, and employ simplifications that are similar to redundancy elimination algorithms (Section 5); they also mention simplifications in intermediate steps of inference that are quite similar to the ideas developed in Section 6. Our work can be viewed as a counterpart of Moral and Cano's ideas for exact algorithms.

Instead of resorting to enumeration techniques, a different approach is to manipulate the values of the joint distribution $p(\mathbf{X})$ directly and optimize the function (2). This approach is inspired by probabilistic logic, an old subject that started with Boole (reviewed by Hansen et al [15]). There are two difficulties with this approach. First, independence relations introduce constraints that are non-linear, a problem that can be circumvented with linearization techniques [1]). Second, the number of elements of $p(\mathbf{X})$ may be huge, a problem that must be coped with column generation techniques [15]. The result is a highly sophisticated algorithm that does not attempt to explore the specificities of credal networks. In this paper we pursue a different path, where we start from enumeration techniques and try to explore the structure of separately specified credal sets.

The only polynomial algorithm for inference with credal networks is the 2U algorithm, which only deals with certain models; we discuss this algorithm in Section 6.

Several methods have been investigated for approximate inference. Tessem has proposed a method that bounds lower and upper probabilities in polytrees with non-binary variables [21]. Cano, Cano and Moral have explored simulated annealing and genetic algorithms in the context of inference [2, 3]; other authors have suggested gradient search and EM-like schemes [7, 23]. In this paper we focus only on *exact* inference.

## 5 REDUNDANCY ELIMINATION

Enumeration methods work by processing every possible selection of vertices in the various sets $K(X_i|\Pi_i)$. This procedure can generate large numbers of potential vertices; for example, the strong extension for the network in Figure 1 can potentially have $2^{20}$ vertices. However, many of these "candidate vertices" may not be actual vertices. More importantly, many of the candidate vertices may be vertices of $K(\mathbf{X})$ but *not* vertices of $K(X_q|\mathbf{X}_E)$. A natural idea then is to discard non-vertices of the strong extension [21]. If we use variable elimination as our base algorithm for inference [9], we must use a redundancy elimination algorithm after the elimination of each variable. It can be proved that this step-by-step redundancy elimination does not affect the final result.

We have implemented redundancy elimination with the extensive form of the CCM transform.[3] We have observed that redundancy elimination fails to provide any computational relief to inferences with strong extensions. The main problem is that the intermediate functions produced by inference algorithms are often defined in high dimensional spaces. Even if we have a large collection of such functions, it is unlikely that many of them will be redundant given the size of the space. We have observed that redundancy elimination rarely removes many candidate vertices when operating in high-dimensional spaces (around 6 di-mensions and larger). We conclude that it is not possible to profit from redundancy elimination in enumeration schemes based on the extensive CCM transform because of the dimensionality of intermediate functions. The main purpose of this paper is to offer a better approach to inference that uses redundancy elimination in an intelligent way (Section 6).

## 6 ALGORITHMS FOR SEPARATELY SPECIFIED STRONG EXTENSIONS

In this section we investigate techniques that can reduce the computational load of inference in separately specified strong extensions. We have argued that separately specified credal sets have a clear semantics and are adequate tools to represent uncertainty. We take that credal networks will typically be associated with separately specified credal sets. We now ask, What techniques can be explored to handle separately specified credal sets?

### 6.1 TERMINAL EVIDENCE

Suppose that we have discarded credal sets that are irrelevant to an inference $\underline{P}(X_q|\mathbf{X}_E)$, using d-separation. The remaining sets form a sub-network that is fully connected. Denote by $\mathbf{X}_{TE}$ the variables in this sub-network that belong to $\mathbf{X}_E$ and that have no children in this sub-network (the "terminal" evidence in the sub-network). The following result considerably simplifies inference by reducing the number of candidate vertices that must be examined.

**Theorem 2** *For every variable $X_i$ in $\mathbf{X}_{TE}$ that has its value fixed to $x_i$, the value of $\underline{P}(X_q|X_E)$ is attained by a distribution that either attains $\overline{p}(X_i = x_i|\Pi_i = \pi_i)$ or $\underline{p}(X_i = x_i|\Pi_i = \pi_i)$ for each value of $\Pi_i$.*

The proof is in Appendix A. This theorem generalizes results of Walley [22, Chap. 8] to multivariate models.

**Example 6** Consider a simple network $X \rightarrow Y \leftarrow Z$, where all variables are ternary and every credal set has 5 vertices. The number of possible vertices for the joint credal set is $5^{11} = 48,828,125$. Using Theorem 2, the number of possible vertices for $K(X|Y)$ is now $5 \times 5 \times 2^9 = 12800$ (assuming no two vertices attain the same lower/upper probabilities).

### 6.2 SEPARABLE VARIABLE ELIMINATION

Consider then that we have applied d-separation and Theorem 2, and arrived at a sub-network for inference. The extensive CCM transform is not appropriate to deal with separately specified credal sets, because transparent variables may have huge numbers of values (Example 5). Clearly, we should keep the original representation $K(X_i|\Pi_i)$. Can we explore this "separable" representation to reduce computational effort? The answer is yes.

---

[3] We used the qconvex and qhull programs for redundancy elimination. Both programs produce exact results (within floating point errors) and are freely available in the world-wide-web.

To simplify our discussion, we assume that inferences are based on *variable elimination* (reviewed in [9]). The results presented here can be adapted to other inference methods. Variable elimination essentially builds a tree of *buckets* and passes messages from the leaves to the root of the tree. Each bucket is responsible for "eliminating" a variable $X_j$. Elimination occurs by selecting all functions that contain $X_j$ (these are the "messages" into the bucket) and summing out $X_j$ from the product of these functions. The result of this operation is a new function that replaces all functions in the bucket and is sent to some other bucket. Note that in a credal network, incoming and outgoing messages are *sets* of functions [5, 7].

Denote by $f_j(\mathbf{X}_j|\mathbf{W}_j)$ one of the functions received by some generic bucket $B$. These functions are unnormalized probability distributions and so we keep the conditioning bar — note that the bucket $B$ can easily discover which variables are in $\mathbf{X}_j$ and in $\mathbf{W}_j$ by examining previous operations. It is important to differentiate between those variables in $\mathbf{W}_j$ that are in the "conditioning" side for all functions received by $B$, and the variables in $\mathbf{W}_j$ that are not. We denote the first group of variables by $\mathbf{Z}_j$ and the second group by $\mathbf{Y}_j$. Consequently, a bucket $B$ receives a number of functions $f_j(\mathbf{X}_j|\mathbf{Y}_j, \mathbf{Z}_j)$, multiplies them, sums out a variable $X$, and sends out a function $f(\mathbf{X}, \mathbf{Y}|\mathbf{Z})$, where $\mathbf{X} = \cup \mathbf{X}_j \backslash X$, $\mathbf{Y} = \cup \mathbf{Y}_j \backslash X$ and $\mathbf{Z} = \cup \mathbf{Z}_j \backslash \{\mathbf{X} \cup \mathbf{Y}\}$.

A crucial result is that set of all functions $f(\mathbf{X}, \mathbf{Y}|\mathbf{Z})$ sent by the bucket $B$ are *separately specified* when we are dealing with separately specified strong extensions. That is, we can operate on any bucket by taking a value of $\mathbf{Z}$ at a time, never concatenating functions with respect to $\mathbf{Z}$. To prove this result, denote by $B_j(\mathbf{X}_j|\mathbf{Y}_j, \mathbf{Z}_j)$ the set of all functions received by bucket $B$ from its $j$th "parent" bucket, and denote by $g(\mathbf{X}, \mathbf{Y}|\mathbf{Z})$ the function $\sum_X \prod_j f_j(\mathbf{X}_j|\mathbf{Y}_j, \mathbf{Z}_j)$.

**Proposition 1** *If for all $j$, $B_j(\mathbf{X}_j|\mathbf{Y}_j, \mathbf{Z}_j)$ is separately specified, then the following sets of functions are identical:*

- $\{g(\mathbf{X}, \mathbf{Y}|\mathbf{Z}) : f_j(\mathbf{X}_j|\mathbf{Y}_j, \mathbf{Z}_j) \in L'_j\}$, *where* $L'_j =$ CONCAT$_{\mathbf{Y}_j, \mathbf{Z}_j}(B_j(\mathbf{X}_j|\mathbf{Y}_j, \mathbf{Z}_j))$.

- CONCAT$_{\mathbf{Z}_j}(\{g(\mathbf{X}, \mathbf{Y}|\mathbf{Z}) : f_j(\mathbf{X}_j|\mathbf{Y}_j, \mathbf{Z}_j) \in L''_j\})$, *where* $L''_j =$ CONCAT$_{\mathbf{Y}_j}(B_j(\mathbf{X}_j|\mathbf{Y}_j, \mathbf{Z}_j))$.

*Proof.* Immediate because any element of the first set can be constructed by operating with the second set, and vice-versa. QED

Note that the first set in Proposition 1 corresponds to brute-force enumeration, while the second set corresponds to keeping separable messages. This proposition suggests a different way to organize computations with separately specified strong extensions: pass messages from bucket to bucket without ever concatenating functions until it becomes necessary to do so. This new organization does not reduce the number of candidate vertices in inference; the main effect of Proposition 1 is to *reduce the dimensionality* of messages sent by buckets during inference. As noted in Section 5, the dimensionality of these functions is the key negative factor in redundancy elimination algorithms. Suppose we run redundancy elimination in these separately specified messages; we obtain the following algorithm:

**Algorithm: Separable variable elimination**

- Run variable elimination, but keep messages $B(\mathbf{X}, \mathbf{Y}|\mathbf{Z})$ sent by buckets as separately specified sets with respect to $\mathbf{Z}$.

- Before a message $B(\mathbf{X}, \mathbf{Y}|\mathbf{Z})$ is sent by a bucket, run redundancy elimination *in each one of the sets* $B(\mathbf{X}, \mathbf{Y}|\mathbf{Z} = \mathbf{z})$ separately, for each value of $\mathbf{Z}$.

The application of redundancy elimination in this fashion has dramatic consequences for inference, as discussed later. Now we focus on the correctness of this algorithm; the next theorem is the main result of the paper, showing the correctness of redundancy elimination over the separately specified messages $B_j(\mathbf{X}_j|\mathbf{Y}_j, \mathbf{Z}_j)$:

**Theorem 3** *If for all $j$, $B_j(\mathbf{X}_j|\mathbf{Y}_j, \mathbf{Z}_j)$ is separately specified, then the following set of functions is equivalent for inference to the sets in Proposition 1:*

CONCAT$_{\mathbf{Z}_j}\left(\text{RE}\{g(\mathbf{X}, \mathbf{Y}|\mathbf{Z}) : f_j(\mathbf{X}_j|\mathbf{Y}_j, \mathbf{Z}_j) \in L''_j\}\right)$,

*where $L''_j$ is defined as in Proposition 1 and* RE $K$ *is the set containing the vertices of $K$ (obtained running redundancy elimination).*

The proof is sketched in Appendix A. Note that a result similar to Theorem 3 must be assumed in order to use the techniques derived by Moral and Cano [4] in the intermediate computations of any inference algorithm.

Separable variable elimination can easily deal with non-binary multi-connected networks, but consider the consequences of the algorithm for polytrees containing no variable with more than $V$ values. Variable elimination can then be organized in such a way that redundancy elimination is never run in dimensions larger than $V$. The gains of separable variable elimination are even more dramatic if $V$ is equal to 2; in this case redundancy elimination is equivalent to a simple inspection for minima and maxima, because every credal set is represented by a single interval (given that functions are probabilities and can be normalized). We have:

**Proposition 2** *For separately specified credal networks composed of polytrees with binary variables, separable variable elimination is an exact polynomial algorithm.*

We have thus obtained the complexity of the 2U algorithm, the only existing method for polynomial inference with credal networks (the 2U algorithm only handles polytrees with binary variables) [13]. The theory of separately specified credal sets developed in this paper leads to a considerable simplification in the derivation of the 2U algorithm, and also makes it possible to state all assumptions actually employed by that algorithm.

We have implemented separable variable elimination on top of the JavaBayes system. We have run several tests with various networks; two examples can illustrate the results. First consider the multi-connected credal network in Figure 1. The query $\underline{P}(F)$ requires the examination of 2393 vertices by separable variable elimination, instead of the $2^{20}$ candidate vertices manipulated by enumeration methods (which could not even finish this inference due to memory exhaustion). As another example, consider the CarStarts network, a polytree with 18 variables produced and distributed by Microsoft Research. We took the original structure, where all variables were binary, and modified the variables Starter and EngineCranks to ternary. We also inserted separately specified credal sets throughout the network, using two vertices per credal set. We then asked the lower probability of the variable Starts. Applying d-separation, we were left with a network with 15 variables, including the ternary ones. Separable variable elimination finished the inference after examining only 3490 vertices of the more than a billion candidate vertices. Again, enumeration methods could not handle the inference.

## 7 CONCLUSION

In this paper, we

- presented a theory of separately specified credal sets, credal networks and strong extensions that can handle existing concepts in the literature and lead to new insights and results;

- proved that inference with polytrees is NP-hard in the presence of credal sets;

- described the difficulties with redundancy elimination in straightforward enumeration;

- introduced new techniques for exact inference with separately specified strong extensions, showing how to use terminal evidence (Theorem 2) and how to perform separable variable elimination;

Separable variable elimination is the first exact algorithm that uses separability of credal sets to an explicit computational advantage. Our tests suggest that the algorithm is viable for small to medium-sized networks, particularly polytrees. The results suggest that inference with semantically meaningful credal networks is a feasible option for uncertainty management.

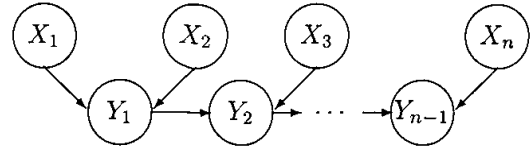

Figure 2: Network that represents SubsetSum problem.

The performance of separable variable elimination depends on the topology of the credal network of interest, and also on the number of values for variables in the network. We believe that the most promising approach for complex networks is to develop branch-and-bound algorithms [16] that take approximations to guide the search for lower/upper probabilities. This strategy seems more effective than general, uninformed methods based on column generation [1] or signomial programming [7, 23]. We are currently investigating and comparing these methods.

## A PROOFS OF THEOREMS

**Theorem 1** We take the NP-complete SubsetSum problem, where we are given a set $S = \{s_1, \ldots, s_n\}$ of non-negative integers and a positive integer $L$; we want to know whether there exists a subset $S' \subseteq S$ such that the sum of elements of $S'$ is exactly $L$ (we are inspired by the work of Lerner and Parr [17]). From a SubsetSum problem, we build a polytree as in Figure 2, with $n$ *value* nodes $X_i$ and $n - 1$ *summation* nodes $Y_i$. Every node has $(1 + \sum_i s_i)$ values (the first value is zero). The summation nodes have two parents and produce the summation of the values of the parents, clipped at $\sum_i s_i$. Suppose that every value node $X_i$ is associated with a credal set with two extreme points: one extreme point places all mass in zero, and the other extreme point places all mass in $s_i$. If we have $\overline{P}(Y_{n-1} = L) = 1$, then it is possible to find $S'$ to produce $L$; otherwise, it is not possible to find such $S'$. Note that if we had a subset $S'$ satisfying the SubsetSum problem (the sum of elements in $S'$ is equal to $L$) then we would verify this solution using a standard polytree propagation (a polynomial procedure). Consequently, inference with such polytrees is NP-complete. QED

**Theorem 2** Using d-separation, we have created a sub-network; the observed leaves of this sub-network are in $\mathbf{X}_{TE}$. Consider one variable $X_i$ in $\mathbf{X}_{TE}$ at a time; it is easy to extend the argument to take all variables in $\mathbf{X}_{TE}$ simultaneously. Take a variable $X_i$ in $\mathbf{X}_{TE}$ that is the child of nodes $\Pi'_i$, possibly containing $X_q$. The set $\Pi'_i$ must contain only the non-observed parents of $X_i$. We can use the *generalized Bayes rule* [22]; the value of $\underline{P}(X_q|X_E)$ is the solution of the following equation in $\lambda$: $\min \sum_{\Pi'_i \setminus X_q} \sum_{X_q} (I(X_q) - \lambda) p(X_i = x_i|\Pi_i) p(X_q, \Pi'_i|\mathbf{X}'_{TE}) = 0$. For every value of $\Pi'_i$, we must select $p(X_i|\Pi_i)$ such that it attains $\underline{p}(X_i = x_i|\Pi_i)$ or $\overline{p}(X_i = x_i|\Pi_i)$. QED

**Theorem 3** *(sketched)* Consider a function $g(\mathbf{X}, \mathbf{Y}|\mathbf{Z} = \mathbf{z})$ that was removed by redundancy elimination at some step of separable variable elimination. We must show that this function cannot be in any extreme point that attains $\underline{P}(X_q|\mathbf{X}_E)$. Note that any extreme point $p(\mathbf{X})$ containing $g(\mathbf{X}, \mathbf{Y}|\mathbf{Z} = \mathbf{z})$ must contain a function $g(\mathbf{X}, \mathbf{Y}|\mathbf{Z})$ of the form $\alpha(\mathbf{Z})g_1(\mathbf{X}, \mathbf{Y}|\mathbf{Z}) + (1-\alpha(\mathbf{Z}))g_2(\mathbf{X}, \mathbf{Y}|\mathbf{Z})$. Now, this function is not a convex combination of $g_1$ and $g_2$ (because $\alpha(\mathbf{Z})$ is a function of $\mathbf{Z}$). But the joint $p(\mathbf{X})$ containing this function must be a convex combination of other joint distributions. To see this, consider the following procedure. First take the functions $[g_1(\mathbf{X}, \mathbf{Y}|\mathbf{z}_1), g_1(\mathbf{X}, \mathbf{Y}|\mathbf{z}_2), \ldots]$ and $[g_2(\mathbf{X}, \mathbf{Y}|\mathbf{z}_1), g_1(\mathbf{X}, \mathbf{Y}|\mathbf{z}_2), \ldots]$, and take their convex combination with mixing factor $\alpha(\mathbf{z}_1)$. This produces a mix *only* for $\mathbf{z}_1$, as the other terms were given by $g_1$. Take also the functions $[g_1(\mathbf{X}, \mathbf{Y}|\mathbf{z}_1), g_2(\mathbf{X}, \mathbf{Y}|\mathbf{z}_2), \ldots]$ and $[g_2(\mathbf{X}, \mathbf{Y}|\mathbf{z}_1), g_2(\mathbf{X}, \mathbf{Y}|\mathbf{z}_2), \ldots]$, and combine them with mixing factor $\alpha(\mathbf{z}_1)$. Now take the convex combination of these two functions, but with various functions that keep $g_1$ and $g_2$ fixed for $\mathbf{z}_2$, $\mathbf{z}_3$, and so on. At the end we have the desired $g(\mathbf{X}, \mathbf{Y}|\mathbf{Z})$ — note that this procedure can only work because we can always select arbitrarily the concatenated functions in a separately specified strong extension. This shows that any removed intermediate function cannot be an extreme point that attains $\underline{P}(X_q|\mathbf{X}_E)$. QED

### Acknowledgments

We thank Marsha Duro from HP Labs, Edson Nery from HP Brasil, and the Instituto de Pesquisas Eldorado for computational facilities in which some of our experiments were conducted. The authors are partially supported by grants from CAPES and CNPq respectively. We also thank the reviewers, particularly the reviewer who brought the important reference [4] to our attention.